\pdfoutput=1
\documentclass[11pt,a4paper]{article}
\usepackage[hyperref]{ranlp2023}
\usepackage{times}
\usepackage{float}
\usepackage{latexsym}
\usepackage{graphicx}
\usepackage{algorithm}      
\usepackage{algorithmic}    
\usepackage{placeins}

\usepackage{microtype}

\aclfinalcopy 

\title{Ngambay-French Neural Machine Translation (sba-Fr)}

\author{Sakayo Toadoum Sari \\
AIMS Senegal \\
  \texttt{tsakayo@aimsammi.org} \\\And
  Angela Fan \\
  Meta AI \\
  \texttt{angelafan@meta.com} 
  \\\And
  Lema Logamou Seknewna\\
 AIMS Senegal \\
  \texttt{seknewna@gmail.com}
  \\}

\date{}

\begin{document}
\maketitle
\begin{abstract}
In Africa, and the world at large, there is an increasing focus on developing Neural Machine Translation (NMT) systems to overcome language barriers. NMT for Low-resource language is particularly compelling as it involves learning with limited labelled data. However, obtaining a well-aligned parallel corpus for low-resource languages can be challenging. The disparity between the technological advancement of a few global languages and the lack of research on NMT for local languages in Chad is striking. End-to-end NMT trials on low-resource Chad languages have not been attempted. Additionally, there is a dearth of online and well-structured data gathering for research in Natural Language Processing, unlike some African languages. However, a guided approach for data gathering can produce bitext data for many Chadian language translation pairs with well-known languages that have ample data. In this project, we created the first sba-Fr Dataset, which is a corpus of Ngambay-to-French translations, and fine-tuned three pre-trained models using this dataset. Our experiments show that the M2M100 model outperforms other models with high BLEU scores on both original and original+synthetic data. The publicly available bitext dataset can be used for research purposes. \footnote{https://github.com/Toadoum/Ngambay-French-Neural-Machine-Translation-sba\_fr\_v1-} 
\end{abstract}

\section{Introduction}
Differential access to information is a pervasive issue in both developed and developing nations, reinforced by physical, social, and economic structures. The problem is especially acute in rural areas, where the lack of communication technology such as the internet can severely limit access to information. Furthermore, automated translation tools face significant challenges in dealing with low-resource language pairs and morphologically rich languages, leading to limited cultural exchange and market integration for certain nations. A major contributor to this problem is the fact that internet research is primarily conducted in languages such as English, French, Spanish, German, etc. resulting in limited data availability for other languages. As a result, Machine Translation (MT) is heavily dependent on parallel text or "bitext," leaving speakers of languages with limited data resources or parallel corpora at a disadvantage when it comes to building MT models \cite{arya}. To make the recent successes of MT systems accessible and inclusive, research efforts should focus on identifying and closing the technological gap between these languages that lack digital or computational data resources. Addressing this gap will require innovative approaches for data collection and processing, as well as the development of new MT models that can effectively operate with limited resources.
The Ngambay language is one of such marginalized and low-resource language facing the challenges of information access and automated translation. As an example of a morphologically rich language, Ngambay encounters significant difficulties in finding adequate translation resources, limiting cultural exchange and economic integration opportunities. The scarcity of internet research conducted in languages like Ngambay further exacerbates this problem, leaving speakers of such languages at a disadvantage in building MT models. Bridging the technological gap for languages with limited digital and computational resources, like Ngambay, is essential to ensure inclusivity and accessibility to the recent successes of MT systems. This research aims to contribute to the advancement of NMT for low-resource languages like Ngambay, making strides toward more equitable access to information and linguistic inclusion.

\section{Related Work}
Machine translation is a crucial subfield of Natural Language Processing (NLP) that utilizes computers to translate natural languages. Recently, end-to-end neural machine translation (NMT) has emerged as the new standard method in practical MT systems, leveraging transformer models with parallel computation and attention mechanism \cite{zhix}. Although NMT models require extensive parallel data, which is typically only available for a limited number of language pairs \cite{surafm}, some research has been conducted on NMT using rare African languages such as Swahili, Hausa, Yoruba, Wolof, Amharic, Bambara, Ghomala, Ewe, Fon, Kinyarwanda, and others. \cite{bonavf} introduced the FFR Dataset, a corpus of Fon-to-French translations, which included the diacritical encoding process and their FFR v1.1 model, trained on the dataset. In their 2020 paper titled "Neural Machine Translation for Extremely Low-Resource African Languages: A Case Study on Bambara," \cite{2011.05284} introduced the pioneering parallel dataset for machine translation of Bambara to and from English and French. This dataset has served as a significant milestone as it has provided the foundation for benchmarking machine translation results involving the Bambara language. The authors extensively address the unique challenges encountered when working with low-resource languages and propose effective strategies to overcome the scarcity of data in low-resource machine translation. Their research sheds light on the potential solutions for improving machine translation in similar linguistic contexts. By tackling the data scarcity issue, \cite{2011.05284}'s work contributes to the advancement of machine translation for under-resourced languages. \cite{davidA} have created a new African news corpus covering 16 languages, including eight that were not part of any existing evaluation dataset. They demonstrated that fine-tuning large pre-trained models with small amounts of high-quality translation data is the most effective strategy for transferring to additional languages and domains. \cite{nekoto2022participatory}, in their paper "Participatory Translations of Oshiwambo", built a resource for language technology development and culture preservation, as well as providing socio-economic opportunities through language preservation. They created a diverse corpus of data spanning topics of cultural importance in the Oshindonga dialect, translated to English, which is the largest parallel corpus for Oshiwambo to-date \cite{nekoto2022participatory}. Other works have also been conducted on African languages, and many of them have websites for data crawling, such as JW300 and BBC. However, there is currently no research related to the Ngambay language or any other local language in Chad, and it is difficult to find websites related to these languages, such as newspapers or other sources, such as JW300.

\section{Ngambay}
Lewis, Simons, and Fennig (2013) reported 896,000 Ngambay speakers in Chad and 57,000 in Cameroon (Wikipedia). According to \cite{book1} J.H. Greenberg's classification in The Languages of Africa places Ngambay in the Nilo-Saharan family, Chari-Nil subfamily, Central Sudanese group, and Bongo-Baguinnian subgroup. Tucker and Bryan classify Ngambay as Bongo-Baguinnian, Sara group. Lakka and Mouroum, closely related to Ngambay, share a fair amount of homogeneity, though they differ in vocabulary and pronunciation. \cite{john} states that Ngambay is related to Western Saras, Kaba, and Laka. Ngambay is spoken in Eastern Logone, Tandjile, Moyen-Chari, Mayo-Kebbi, and Chari-Baguirmi prefectures. It is used as a lingua franca by other ethnic groups. In 1993, 812,003 Ngambay lived in Chad, with at least half in Logone Occidental. The Ngambay people call their language "tàr Ngàmbáí" or "tà Ngàmbáí". Protestant priests and missionaries helped many Ngambay speakers learn to read and write. They translated the New Testament and Bible into Ngambay, titled "Testament ge cigi" and "Maktub ge to qe kemee" respectively. Ngambay hymns include "Pa kula ronduba do Mbaidombaije'g". It is worth noting that a monthly evangelical magazine called Dannasur was published for several decades until its discontinuation in 1995, or possibly more recently.

However, it is regrettable that the transcription of Ngambay has not taken into account its distinctive feature of tones. Several studies have already been conducted on this language, including the work of Charles Vandame (Archbishop of N'Djamena before) titled The Ngambay-Moundou, which was published in 1963 \cite{book1}.

\section{Problem of Education}
The economic difficulties of recent years have had a significant impact on the education sector of Chad, leading to stagnation or even a decline in the quality and effectiveness of the education system. School infrastructure has deteriorated rapidly, and there is a lack of motivated and qualified staff, with illiteracy remaining prevalent and gender disparities showing no signs of improvement. Although the primary school enrollment rate is relatively high at 86.85\%, only 41.32\% of students complete primary school. When compared with Niger, a neighbouring African country facing similar challenges, the data is disappointing, with Niger having a primary school enrollment rate of 73.43\% and nearly 72\% of students completing primary school. A recent sectoral analysis of the Chadian education system highlights several deficiencies, including low enrollment rates, a lack of textbooks and inadequate classroom equipment, unqualified teachers, and limited access to higher education. Therefore, several changes are necessary to improve education in Chad. The PAQEPP (Projet d'amélioration de la qualité de l'éducation par une gestion de proximité) project, funded by the French Development Agency, aims to address these issues, involving 50 schools in Moundou and N'Djamena. The project was scheduled to run for four years, from 2017 to 2021, and involved more than 700 teachers and nearly 55,000 students. However, due to the global health crisis (COVID-19), the project has been extended until 2023.\\

One possible solution to address such problems is the development of efficient Machine Translation Models that can be deployed on edge devices to help overcome language barriers, as many people face difficulties in accessing education. Creating high-quality datasets for research in NMT is crucial for building these models.
\section{Data creation}
In data creation, we utilized two sources. The first source was \href{https://morkegbooks.com/Services/World/Languages/SaraBagirmi/#title}{\textit{The Sara Bagirmi Languages Project}} which provided us with the fifth edition (2015) of the Ngambaye to French dictionary in PDF format. However, due to the complexity of performing web scraping on a PDF, we manually created a parallel corpus of 1,176 sentences with short to medium lengths from the most commonly used sentences in daily life using a Google form. The second source was \href{https://chop.bible.com/fr/}{\textit{YouVersion Bible}}, an online Bible translated into multiple languages, including Ngambay. Using R programming, we performed web scraping on the website, but the Ngambay translation did not include all the verses like the French version. We extracted up to 34,647 sentences, but there were various grammatical errors, incorrect and incomplete translations, and inconsistencies. To ensure the quality of the data after crawling, we gave the dataset to native speakers of Ngambay and other linguists, including the Association of People translating the Bible from French to Ngambay in Chad, to check for problematic translations, misspellings, and duplicated sentences following \citet{partic}. After quality control, we combined the two bitext datasets, dropped inconsistent and incomplete translations, and ended up with 33,073 sentences for use in this project.

The morphological characteristics of a language can have a significant impact on its sentence structure and complexity. Our analysis revealed that the Ngambay language has a relatively simple morphology compared to French, which contributes to shorter sentences and fewer words. In contrast, French has a highly inflected morphology, resulting in longer and more complex sentences with a larger vocabulary. These differences in morphology pose a challenge for Machine Translation systems, as they must be trained on parallel texts that are aligned at the sentence and word levels. Given the complexity of French and the simplicity of Ngambay, it is essential to develop effective strategies for handling the morphological variations in each language when building MT models. By understanding the unique features of each language, we can improve the accuracy and effectiveness of MT systems for languages with varying levels of complexity.



\subsection{Data Split}
Splitting our bitext data into training, validation, and test sets using a 20\% split size is a common ML practice for creating reliable, precise, and generalizable models. After splitting, our sets had 21,166, 6,615, and 5,292 sentences respectively for train, validation and test. We used the Python package jsonlines\footnote{https://jsonlines.readthedocs.io/en/latest/} to convert our CSV files to JSON format to match Hugging Face's pre-trained models.

\section{Models and Methods}
We have used three transformer-based language models in our experiments: MT5 \cite{xue}, ByT5 \cite{linting}, and M2M100 \cite{angela}. Transformers are a type of neural network architecture that has become popular in NLP since 2017 \cite{ashish}. They are used in many cutting-edge NLP applications. Unlike RNNs, transformers use a self-attention mechanism to weigh input sequence importance when making predictions. The transformer architecture consists of an encoder and decoder, which can be trained for NLP tasks such as machine translation, text classification, and language modelling. The encoder produces hidden representations from the input sequence, and the decoder uses them to generate the output sequence \cite{ashish}.


\subsection{M2M100}
M2M100 is a large multilingual machine translation model proposed by \cite{angela}. It uses a shared representation space and a pivot language to enable translations between 100 languages, including low-resource and non-Indo-European languages. The model outperforms previous multilingual models and achieves state-of-the-art results on various translation benchmarks\cite{angela}.\\

\subsection{ByT5}
ByT5 is a byte-to-byte transformer model introduced by \cite{linting}. It operates at the byte level, eliminating the need for tokenization and making it suitable for languages with complex scripts or non-standard formatting. ByT5 outperforms existing token-based models on benchmark datasets, including those with low-resource languages \cite{linting}.

\subsection{MT5}
MT5 is a massively multilingual pre-trained text-to-text transformer proposed by \cite{xue}. It is trained on a large corpus of text in over 100 languages and can directly translate between any pair of languages without relying on English as an intermediate step. The text-to-text approach and diverse training tasks contribute to its versatility and performance \cite{xue}.\\

Fine-tuning pre-trained models on a new low-resource language like Ngambay requires careful consideration of the available data and the best approach to utilizing it. As noted by \cite{davidA}, one effective way to fine-tune pre-trained models is to follow a process. It is essential to select a target language that is represented in all the pre-trained models. In this case, we chose Swahili (sw) as our target language since it is a commonly used language that is present in most pre-trained models. This allows us to leverage the existing knowledge contained in the pre-trained models and adapt it to the new African language \cite{davidA}.
\subsection{Hardware and Schedule}
Our models were trained on a single machine equipped with 2 NVIDIA T4 GPUs, 32 vCPUs, and 120 GB of RAM. During the training process, optimization steps for M2M100, ByT5, and MT5 took an average of 5 seconds, 2 seconds, and 4 seconds, respectively, based on the pre-trained models and hyperparameter described in the section \ref{subsection:hyper}. We trained our models for a total of 133,080 optimization steps. The M2M100 model was trained for 1 day, 15:02:53.55, ByT5 for 1 day, 0:56:06.98, and MT5 for 20:46:36.98.

\subsection{Performance Evaluation Metrics and Hyperparameters}
\label{subsection:hyper}
In this project, we utilized BLEU as a means of automatically evaluating machine translation. BLEU evaluates the adequacy of machine translation by analyzing word precision, as well as the fluency of the translation by calculating n-gram precisions. This method returns a score within a range of [0, 1] or on a [0, 100] scale. We specifically implemented SacreBLEU, which provides dataset scores instead of segment scores. A higher score indicates a translation that is closer to the reference \cite{kish}:\\

Using the HuggingFace transformer tool, we fine-tuned pre-trained models with settings that included a learning rate of 5e-5, a batch size of 5, maximum source and target lengths of 200, a beam size of 10, and a total of 60 epochs.
\section{Results and Discussion}
This section will detail our training process, specifically discussing the data augmentation method we used to enhance the performance of our pre-trained models. Our source language is French (Fr), while the target language is Ngambay (sba).\\

Our experiment aimed to identify and select the model that performed best among the pre-trained models when trained on the original bitext data, then use the selected model to generate synthetic data. Of the three pre-trained models we fine-tuned, M2M100 achieved the highest Evaluation BLEU score of 33.06, followed by ByT5 with a score of 28.447 when trained on a sample of 21,166, as shown in Table ~\ref{tab:orig}. This can be attributed to the fact that M2M100 is a multilingual model trained on a diverse set of parallel corpora from 100 languages, including news articles, subtitles, and other publicly available texts. It employs a shared encoder-decoder architecture that can be fine-tuned for specific language pairs and integrates multiple techniques to improve performance \cite{angela}.



\begin{table*}[!htbp]
\centering
\begin{tabular}{lccc}
\hline
\textbf{Models} & \textbf{M2M100} &\textbf{ByT5} &\textbf{MT5} \\
\hline
Eval BLEU  & 33.06 & 28.447 & 22.12 \\
Predict BLEU & 32.6016 & 32.6016& 22.0481\\
Eval loss & 1.7661 & 0.5152& 1.0874\\ 
Train sample & 1166 & 24366& 21166\\
Train runtime & 1 day, 15:02:53.55 &1 day, 0:56:06.98 & 20:46:36.98  \\\hline
\end{tabular}
\caption{Result of Fine-tuning  M2M100, ByT5, and, MT5 using original Dataset.}
\label{tab:orig}
\end{table*}

\subsection{Data Augmentation using French monolingual data}
\label{subsection:augment}
In their 2016 paper, \cite{rico} proposed a method to enhance NMT models with available monolingual data for many languages. The two-step process involves training a language model on the bitext data and then using it to generate synthetic parallel sentences for the NMT model by translating the monolingual sentences into the target language \cite{rico}. \cite{atnafu} proposed Source-side Monolingual Data Injection (SMDI) to enhance low-resource NMT systems. A language model is trained on a parallel corpus and used to generate synthetic parallel sentences by translating the monolingual sentences into the target language. Evaluations on several low-resource language pairs showed that SMDI consistently improved NMT system quality \cite{atnafu}.

We are tackling a low-resource language with little in-domain data for Neural Machine Translation. Thus, we use a method similar to \cite{rico}. To generate synthetic parallel data for Ngambay-French translation we have used the fra\_news\_2022\_100K-sentences.txt dataset from the Leipzig Corpora Collection/Deutscher Wortschatz, containing 100,000 sentences related to 2022 news (politics, sport, entertainment, etc.) because no monoligual Ngambay data exists, unless in hard copy, hence, input (Fr) monolingual source-side. We create synthetic bitext data from French monolingual data. We split the monolingual data into sentences, and perform noisy translation to Ngambay then combine the translated sentences to form a synthetic bitext corpus.

\begin{algorithm}[h!]
\caption{Generating synthetic bitext data \& training}
\begin{algorithmic} 
\REQUIRE 
\begin{itemize}
\item Original bitext dataset: $sba-Fr $
\item French Monolingual dataset: $Fr_m$
\item Target synthetic dataset: $sba_{synth}$
\item Synthetic bitext dataset: $sba_{synth}-Fr_m$
\item Languages: Fr, sba
\item Translation model: NMT Fr → sba
\end{itemize}
\ENSURE 
\begin{itemize}
    \item Train NMT on $sba-Fr$
    \item split $Fr_m$ into sentences 
    \item generate synthetic $sba_{synth}$ by translating $Fr_m$ sentences  using trained and saved NMT
\item Combine sentences from $Fr_m$ and $sba_{synth}$ to create $sba_{synth}-Fr_m$
\item Add $sba-Fr$ and $sba_{synth}-Fr_m$ to create new bitext data
\item Retrain the model using the new bitext data.
\end{itemize}
\end{algorithmic}
\end{algorithm}
In machine translation, a model is typically trained on original bitext data, and then utilized to translate a set of monolingual source sentences into the target language. This process generates pseudo-parallel training data, also known as synthetic data. The synthetic data is subsequently combined with the authentic parallel data to train and improved the model, following the self-training concept introduced by \cite{junxian}. This involves training a model on labelled data and using it to generate pseudo-labelled data, which is then added to the training set to enhance the model's performance \cite{junxian}.

We used French monolingual data to generate translations for Ngambay. We combined these to create synthetic bitext data (see section ~\ref{subsection:augment}). Training our models on both the original and synthetic data increased the M2M100 and ByT5 model's Evaluation BLEU score by more than 11 points compared to the original data alone. The MT5 model's Evaluation BLEU score increased by more than 2 points compared to the original dataset. This result is consistent with \citet{atnafu}, who used target monolingual data in self-training experiments. Table ~\ref{tab:orig+synt} shows that M2M100 outperforms the other two models with original and original + synthetic data. \cite{agostinho} with their work "Findings from the Bambara - French Machine Translation Competition (BFMT 2023)" have used Cyclic backtranslation, aims to enhance the model's learning by utilizing both the training dataset and a monolingual dataset. At each step $k$, they encourage the Machine Translation (MT) model for each direction to learn from a combination of the original training dataset, sentences generated synthetically, and sentences generated by the MT model of the opposite direction from the previous step. This approach allows the model to benefit from the diverse data sources, leading to improved performance and robustness. They have also used M2M100 model \cite{angela} as their starting point due to its outstanding performance, achieving the highest scores.  \cite{davidA} demonstrated this in their project entitled ``A Few Thousand Translations Go A Long Way!", they created an African news corpus with 16 languages, including 8 not in any existing evaluation dataset. M2M100 adapts faster than ByT5, and in most cases, it outperforms the other models and this have been confirmed by \cite{2207.04672}'s results. The M2M100 model is capable of translating between 100 languages in a many-to-many manner, which means it can translate any language pair among the 100 supported languages. The model is trained using a novel approach called Cyclic Backtranslation, which enables the model to learn from both the original training dataset and a synthetic dataset generated through translation of monolingual dataset. By leveraging a large amount of multilingual data, the M2M100 model demonstrates significant improvements in translation quality for various language pairs. Hence, it consistently delivers superior results in most cases.

\begin{table*}[!htbp]
\centering
\begin{tabular}{lccc}
\hline
\textbf{Models} & \textbf{M2M100} &\textbf{ByT5} &\textbf{MT5} \\
\hline
Eval BLEU  & 53.1034 & 43.0504 & 24.6858 \\
Predict BLEU & 52.6012 & 42.52518& 24.4494\\
Eval loss & 1.1236 & 0.2801& 0.9246\\ 
Train sample  & 24366 & 24366& 24366\\
Train runtime & 1 day, 12:00:14.38 &1 day, 13:36:28.15 & 1 day, 3:10:09.87  \\\hline
\end{tabular}
\caption{Result of Fine-tuning  M2M100, ByT5, and, MT5 using original + synthetic Dataset.}
\label{tab:orig+synt}
\end{table*}

\section{Conclusion}
The primary objective of this study is to demonstrate the possibility of gathering data on Chadian languages, similar to how other African countries do, and utilizing this data to develop a Machine Translation (MT) system. Specifically, the aim is to establish an MT system for the Ngambay language as an example for other Chadian languages. By doing so, we hope to set a benchmark for the accuracy of Chadian MT systems. To achieve this goal, we constructed the first bitext dataset for Ngambay-French and fine-tuned three transformer-based models (M2M100, ByT5, and MT5). Our experimental results indicate that M2M100 outperforms the other models and that monolingual source-side can enhance the performance of all models. We believe that such MT system can be integrated into electronic devices to overcome language barriers. However, this work has limitations that future studies can address


\section{Limitations}
Challenges exist in developing Neural Machine Translation (NMT) systems for low-resource languages in Chad. Obtaining a well-aligned parallel corpus is difficult, leading to inadequate training in translation models. Furthermore, technological advancement in NMT focuses on global languages, leaving a research gap for local languages in Chad. Consequently, end-to-end NMT trials for low-resource Chad languages have not been conducted. Online and structured data gathering for NLP research in Chadian languages is limited, making it hard to acquire enough data for successful NMT model training. A guided approach was used with languages having abundant data, but this may not capture the local languages' complexities, potentially affecting model performance. The M2M100 model's generalization to other low-resource Chadian languages is uncertain. Biases in the sba-Fr Dataset used in the project could affect the model's accuracy and practicality.

\section{Future Work}
To address the limitations of our current study, future research can focus on several aspects. Firstly, our dataset predominantly originates from the bible, which may introduce biased religious references. To mitigate this bias, researchers can collect more diverse and general text data for the Ngambay language.

Additionally, exploring advanced techniques like circular Back-translation using monolingual target source-side and Meta-Learning for Few-Shot NMT Adaptation, as proposed \cite{rico} and \cite{kim-etal-2019-effective} respectively, could lead to enhancements in both the dataset quality and the overall performance of the machine translation (MT) system. These techniques have shown promise in improving MT systems by leveraging additional data and adapting to low-resource languages like Ngambay more effectively.

\section{Acknowledgments}
We express our gratitude to the African Institute for Mathematical Sciences (AIMS) with the program African Master's of Machine Intelligence (AMMI) for providing us with high-quality machine-learning training and for supporting us throughout this project. We also extend our appreciation to Google for providing us with a Google Cloud Platform (GCP) grant that allowed us to run our experiments. Special thanks go to the AMMI staff for their assistance and support. Many thanks to Chris Emezue and Lyse Naomi Wamba for the proofreading and useful comments.

\bibliographystyle{acl_natbib}
\bibliography{ranlp2023}

\begin{thebibliography}{21}
\expandafter\ifx\csname natexlab\endcsname\relax\def\natexlab#1{#1}\fi

\bibitem[{Adelani et~al.(2022)Adelani, Alabi, Fan, Kreutzer, Shen, Reid,
  Ruiter, Klakow, Nabende, Chang, Gwadabe, Sackey, Dossou, Emezue, Leong,
  Beukman, Muhammad, Jarso, Yousuf, Niyongabo~Rubungo, Hacheme, Wairagala,
  Nasir, Ajibade, Ajayi, Gitau, Abbott, Ahmed, Ochieng, Aremu, Ogayo, Mukiibi,
  Ouoba~Kabore, Kalipe, Mbaye, Tapo, Memdjokam~Koagne, Munkoh-Buabeng, Wagner,
  Abdulmumin, Awokoya, Buzaaba, Sibanda, Bukula, and Manthalu}]{davidA}
David Adelani, Jesujoba Alabi, Angela Fan, Julia Kreutzer, Xiaoyu Shen, Machel
  Reid, Dana Ruiter, Dietrich Klakow, Peter Nabende, Ernie Chang, Tajuddeen
  Gwadabe, Freshia Sackey, Bonaventure F.~P. Dossou, Chris Emezue, Colin Leong,
  Michael Beukman, Shamsuddeen Muhammad, Guyo Jarso, Oreen Yousuf, Andre
  Niyongabo~Rubungo, Gilles Hacheme, Eric~Peter Wairagala, Muhammad~Umair
  Nasir, Benjamin Ajibade, Tunde Ajayi, Yvonne Gitau, Jade Abbott, Mohamed
  Ahmed, Millicent Ochieng, Anuoluwapo Aremu, Perez Ogayo, Jonathan Mukiibi,
  Fatoumata Ouoba~Kabore, Godson Kalipe, Derguene Mbaye, Allahsera~Auguste
  Tapo, Victoire Memdjokam~Koagne, Edwin Munkoh-Buabeng, Valencia Wagner, Idris
  Abdulmumin, Ayodele Awokoya, Happy Buzaaba, Blessing Sibanda, Andiswa Bukula,
  and Sam Manthalu. 2022.
\newblock \href {https://doi.org/10.18653/v1/2022.naacl-main.223} {A few
  thousand translations go a long way! leveraging pre-trained models for
  {A}frican news translation}.
\newblock In \emph{Proceedings of the 2022 Conference of the North American
  Chapter of the Association for Computational Linguistics: Human Language
  Technologies}, pages 3053--3070, Seattle, United States. Association for
  Computational Linguistics.

\bibitem[{Agostinho Da~Silva et~al.(2023)Agostinho Da~Silva, Ajayi, Antonov,
  Azazia~Kamate, Coulibaly, Del~Rio, Diarra, Diarra, Emezue, Hamilcaro, Homan,
  Most, Mwatukange, Ohue, Pham, Sako, Samb, Sy, Weerasooriya, Zahidi, and
  Luger}]{agostinho}
Ninoh Agostinho Da~Silva, Tunde~Oluwaseyi Ajayi, Alexander Antonov, Panga
  Azazia~Kamate, Moussa Coulibaly, Mason Del~Rio, Yacouba Diarra, Sebastian
  Diarra, Chris Emezue, Joel Hamilcaro, Christopher~M. Homan, Alexander Most,
  Joseph Mwatukange, Peter Ohue, Michael Pham, Abdoulaye Sako, Sokhar Samb,
  Yaya Sy, Tharindu~Cyril Weerasooriya, Yacine Zahidi, and Sarah Luger. 2023.
\newblock \href {https://aclanthology.org/2023.loresmt-1.9} {Findings from the
  {B}ambara - {F}rench machine translation competition ({BFMT} 2023)}.
\newblock In \emph{Proceedings of the The Sixth Workshop on Technologies for
  Machine Translation of Low-Resource Languages (LoResMT 2023)}, pages
  110--122, Dubrovnik, Croatia. Association for Computational Linguistics.

\bibitem[{Emezue and Dossou(2020)}]{bonavf}
Chris~Chinenye Emezue and Femi Pancrace~Bonaventure Dossou. 2020.
\newblock \href {https://doi.org/10.18653/v1/2020.winlp-1.21} {{FFR} v1.1:
  {F}on-{F}rench neural machine translation}.
\newblock In \emph{Proceedings of the The Fourth Widening Natural Language
  Processing Workshop}, pages 83--87, Seattle, USA. Association for
  Computational Linguistics.

\bibitem[{Fan et~al.(2021)Fan, Bhosale, Schwenk, Ma, El-Kishky, Goyal, Baines,
  Celebi, Wenzek, Chaudhary et~al.}]{angela}
Angela Fan, Shruti Bhosale, Holger Schwenk, Zhiyi Ma, Ahmed El-Kishky,
  Siddharth Goyal, Mandeep Baines, Onur Celebi, Guillaume Wenzek, Vishrav
  Chaudhary, et~al. 2021.
\newblock Beyond english-centric multilingual machine translation.
\newblock \emph{The Journal of Machine Learning Research}, 22(1):4839--4886.

\bibitem[{He et~al.(2020)He, Gu, Shen, and Ranzato}]{junxian}
Junxian He, Jiatao Gu, Jiajun Shen, and Marc'Aurelio Ranzato. 2020.
\newblock \href {https://openreview.net/forum?id=SJgdnAVKDH} {Revisiting
  self-training for neural sequence generation}.
\newblock In \emph{International Conference on Learning Representations}.

\bibitem[{John(2012)}]{john}
M.~Keegan John. 2012.
\newblock The sara bagirmi languages project.
\newblock \emph{Morkegbooks}.
\newblock Accessed: August 2022.

\bibitem[{Kim et~al.(2019)Kim, Gao, and Ney}]{kim-etal-2019-effective}
Yunsu Kim, Yingbo Gao, and Hermann Ney. 2019.
\newblock \href {https://doi.org/10.18653/v1/P19-1120} {Effective cross-lingual
  transfer of neural machine translation models without shared vocabularies}.
\newblock In \emph{Proceedings of the 57th Annual Meeting of the Association
  for Computational Linguistics}, pages 1246--1257, Florence, Italy.
  Association for Computational Linguistics.

\bibitem[{McCarthy(2017)}]{arya}
Arya McCarthy. 2017.
\newblock The new digital divide: Language is the impediment to information
  acccess.
\newblock \emph{HILLTOPICS}.
\newblock Accessed: August 2022.

\bibitem[{Ndjérassem(2000)}]{book1}
Mbai-Yelmia~Ngabo Ndjérassem. 2000.
\newblock \emph{Phonologie du Ngambai, parler de Benoye (Tchad)}, volume 12/13
  of \emph{University of Leipzig papers on Africa (ULPA): languages and
  literatures series}.
\newblock Institut für Afrikanistik, Univ. Leipzig, Leipzig.
\newblock Includes bibliographical references (p. 71-74).

\bibitem[{Nekoto et~al.(2022)Nekoto, Kreutzer, Rajab, Ochieng, and
  Abbott}]{nekoto2022participatory}
Wilhelmina Nekoto, Julia Kreutzer, Jenalea Rajab, Millicent Ochieng, and Jade
  Abbott. 2022.
\newblock \href
  {https://www.microsoft.com/en-us/research/publication/participatory-translations-of-oshiwambo-towards-sustainable-culture-preservation-with-language-technology/}
  {Participatory translations of oshiwambo: Towards sustainable culture
  preservation with language technology}.
\newblock \emph{AfricaNLP}.

\bibitem[{Nekoto et~al.(2020)Nekoto, Marivate, Matsila, Fasubaa, Fagbohungbe,
  Akinola, Muhammad, Kabongo~Kabenamualu, Osei, Sackey, Niyongabo, Macharm,
  Ogayo, Ahia, Berhe, Adeyemi, Mokgesi-Selinga, Okegbemi, Martinus, Tajudeen,
  Degila, Ogueji, Siminyu, Kreutzer, Webster, Ali, Abbott, Orife, Ezeani,
  Dangana, Kamper, Elsahar, Duru, Kioko, Espoir, van Biljon, Whitenack,
  Onyefuluchi, Emezue, Dossou, Sibanda, Bassey, Olabiyi, Ramkilowan, {\"O}ktem,
  Akinfaderin, and Bashir}]{partic}
Wilhelmina Nekoto, Vukosi Marivate, Tshinondiwa Matsila, Timi Fasubaa, Taiwo
  Fagbohungbe, Solomon~Oluwole Akinola, Shamsuddeen Muhammad, Salomon
  Kabongo~Kabenamualu, Salomey Osei, Freshia Sackey, Rubungo~Andre Niyongabo,
  Ricky Macharm, Perez Ogayo, Orevaoghene Ahia, Musie~Meressa Berhe, Mofetoluwa
  Adeyemi, Masabata Mokgesi-Selinga, Lawrence Okegbemi, Laura Martinus,
  Kolawole Tajudeen, Kevin Degila, Kelechi Ogueji, Kathleen Siminyu, Julia
  Kreutzer, Jason Webster, Jamiil~Toure Ali, Jade Abbott, Iroro Orife, Ignatius
  Ezeani, Idris~Abdulkadir Dangana, Herman Kamper, Hady Elsahar, Goodness Duru,
  Ghollah Kioko, Murhabazi Espoir, Elan van Biljon, Daniel Whitenack,
  Christopher Onyefuluchi, Chris~Chinenye Emezue, Bonaventure F.~P. Dossou,
  Blessing Sibanda, Blessing Bassey, Ayodele Olabiyi, Arshath Ramkilowan, Alp
  {\"O}ktem, Adewale Akinfaderin, and Abdallah Bashir. 2020.
\newblock \href {https://doi.org/10.18653/v1/2020.findings-emnlp.195}
  {Participatory research for low-resourced machine translation: A case study
  in {A}frican languages}.
\newblock In \emph{Findings of the Association for Computational Linguistics:
  EMNLP 2020}, pages 2144--2160, Online. Association for Computational
  Linguistics.

\bibitem[{Papineni et~al.(2002)Papineni, Roukos, Ward, and Zhu}]{kish}
Kishore Papineni, Salim Roukos, Todd Ward, and Wei-Jing Zhu. 2002.
\newblock \href {https://doi.org/10.3115/1073083.1073135} {{B}leu: a method for
  automatic evaluation of machine translation}.
\newblock In \emph{Proceedings of the 40th Annual Meeting of the Association
  for Computational Linguistics}, pages 311--318, Philadelphia, Pennsylvania,
  USA. Association for Computational Linguistics.

\bibitem[{Sennrich et~al.(2016)Sennrich, Haddow, and Birch}]{rico}
Rico Sennrich, Barry Haddow, and Alexandra Birch. 2016.
\newblock \href {https://doi.org/10.18653/v1/P16-1009} {Improving neural
  machine translation models with monolingual data}.
\newblock In \emph{Proceedings of the 54th Annual Meeting of the Association
  for Computational Linguistics (Volume 1: Long Papers)}, pages 86--96, Berlin,
  Germany. Association for Computational Linguistics.

\bibitem[{Surafel et~al.(2018)Surafel, Marcello, Matteo, and Marco}]{surafm}
M.~Lakew Surafel, Federico Marcello, Negri Matteo, and Turchi Marco. 2018.
\newblock Multilingual neural machine translation for low-resource languages.
\newblock \emph{Emerging Topics at the Fouth Italian Conference on
  Computational Linguistics (Part 1)}, pages 11--25.

\bibitem[{Tapo et~al.(2020)Tapo, Coulibaly, Diarra, Homan, Kreutzer, Luger,
  Nagashima, Zampieri, and Leventhal}]{2011.05284}
Allahsera~Auguste Tapo, Bakary Coulibaly, Sébastien Diarra, Christopher Homan,
  Julia Kreutzer, Sarah Luger, Arthur Nagashima, Marcos Zampieri, and Michael
  Leventhal. 2020.
\newblock \href {http://arxiv.org/abs/arXiv:2011.05284} {Neural machine
  translation for extremely low-resource african languages: A case study on
  bambara}.

\bibitem[{Team et~al.(2022)Team, Costa-jussà, Cross, Çelebi, Elbayad,
  Heafield, Heffernan, Kalbassi, Lam, Licht, Maillard, Sun, Wang, Wenzek,
  Youngblood, Akula, Barrault, Gonzalez, Hansanti, Hoffman, Jarrett, Sadagopan,
  Rowe, Spruit, Tran, Andrews, Ayan, Bhosale, Edunov, Fan, Gao, Goswami,
  Guzmán, Koehn, Mourachko, Ropers, Saleem, Schwenk, and Wang}]{2207.04672}
NLLB Team, Marta~R. Costa-jussà, James Cross, Onur Çelebi, Maha Elbayad,
  Kenneth Heafield, Kevin Heffernan, Elahe Kalbassi, Janice Lam, Daniel Licht,
  Jean Maillard, Anna Sun, Skyler Wang, Guillaume Wenzek, Al~Youngblood, Bapi
  Akula, Loic Barrault, Gabriel~Mejia Gonzalez, Prangthip Hansanti, John
  Hoffman, Semarley Jarrett, Kaushik~Ram Sadagopan, Dirk Rowe, Shannon Spruit,
  Chau Tran, Pierre Andrews, Necip~Fazil Ayan, Shruti Bhosale, Sergey Edunov,
  Angela Fan, Cynthia Gao, Vedanuj Goswami, Francisco Guzmán, Philipp Koehn,
  Alexandre Mourachko, Christophe Ropers, Safiyyah Saleem, Holger Schwenk, and
  Jeff Wang. 2022.
\newblock \href {http://arxiv.org/abs/arXiv:2207.04672} {No language left
  behind: Scaling human-centered machine translation}.

\bibitem[{Tonja et~al.(2023)Tonja, Kolesnikova, Gelbukh, and Sidorov}]{atnafu}
Atnafu~Lambebo Tonja, Olga Kolesnikova, Alexander Gelbukh, and Grigori Sidorov.
  2023.
\newblock \href {https://doi.org/10.3390/app13021201} {Low-resource neural
  machine translation improvement using source-side monolingual data}.
\newblock \emph{Applied Sciences}, 13(2).

\bibitem[{Vaswani et~al.(2017)Vaswani, Shazeer, Parmar, Uszkoreit, Jones,
  Gomez, Kaiser, and Polosukhin}]{ashish}
Ashish Vaswani, Noam Shazeer, Niki Parmar, Jakob Uszkoreit, Llion Jones,
  Aidan~N Gomez, \L~ukasz Kaiser, and Illia Polosukhin. 2017.
\newblock \href
  {https://proceedings.neurips.cc/paper_files/paper/2017/file/3f5ee243547dee91fbd053c1c4a845aa-Paper.pdf}
  {Attention is all you need}.
\newblock In \emph{Advances in Neural Information Processing Systems},
  volume~30. Curran Associates, Inc.

\bibitem[{Xue et~al.(2022)Xue, Barua, Constant, Al-Rfou, Narang, Kale, Roberts,
  and Raffel}]{linting}
Linting Xue, Aditya Barua, Noah Constant, Rami Al-Rfou, Sharan Narang, Mihir
  Kale, Adam Roberts, and Colin Raffel. 2022.
\newblock \href {https://doi.org/10.1162/tacl_a_00461} {{B}y{T}5: Towards a
  token-free future with pre-trained byte-to-byte models}.
\newblock \emph{Transactions of the Association for Computational Linguistics},
  10:291--306.

\bibitem[{Xue et~al.(2021)Xue, Constant, Roberts, Kale, Al-Rfou, Siddhant,
  Barua, and Raffel}]{xue}
Linting Xue, Noah Constant, Adam Roberts, Mihir Kale, Rami Al-Rfou, Aditya
  Siddhant, Aditya Barua, and Colin Raffel. 2021.
\newblock \href {https://doi.org/10.18653/v1/2021.naacl-main.41} {m{T}5: A
  massively multilingual pre-trained text-to-text transformer}.
\newblock In \emph{Proceedings of the 2021 Conference of the North American
  Chapter of the Association for Computational Linguistics: Human Language
  Technologies}, pages 483--498, Online. Association for Computational
  Linguistics.

\bibitem[{Zhixing et~al.(2020)Zhixing, Shuo, Zonghan, Gang, Xuancheng, Maosong,
  and Yang}]{zhix}
Tan Zhixing, Wang Shuo, Yang Zonghan, Chen Gang, Huang Xuancheng, Sun Maosong,
  and Liu Yang. 2020.
\newblock Neural machine translation: A review of methods, resources, and
  tools.
\newblock \emph{AI Open}, 1:5--21.

\end{thebibliography}


\end{document}